\documentclass[10pt,twocolumn,letterpaper]{article}

\usepackage{cvpr}              

\usepackage{appendix}
\usepackage{float}
\usepackage{booktabs}
\usepackage{pifont}
\usepackage{multirow}
\usepackage{makecell}

\usepackage{algorithm}
\usepackage{algpseudocode}
\usepackage{amsmath}
\usepackage{amssymb}

\newcommand{\X}{\mathcal{X}}       
\newcommand{\Xt}{\tilde{\mathcal{X}}} 
\newcommand{\Xh}{\hat{\mathcal{X}}}   
\newcommand{\K}{\mathcal{K}}       

\definecolor{cvprblue}{rgb}{0.21,0.49,0.74}
\usepackage[pagebackref,breaklinks,colorlinks,allcolors=cvprblue]{hyperref}

\usepackage[table]{xcolor}

\def\paperID{798} 
\def\confName{CVPR}
\def\confYear{2026}

\title{NOVA: Sparse Control, Dense Synthesis for Pair-Free Video Editing}

\author{
    {Tianlin Pan}$^{1,2}$ \quad {Jiayi Dai}$^{1}$ \quad {Chenpu Yuan}$^{1,2}$ \quad {Zhengyao Lv}$^{4}$ \quad {Binxin Yang}$^{3}$ \\ \quad {Hubery Yin}$^{3}$ \quad {Chen Li}$^{3\dag}$ \quad {Jing Lyu}$^{3}$ \quad {Caifeng Shan}$^{1}$\quad {Chenyang Si}$^{1\dag}$ \\ \\
     $^1$Nanjing University \quad $^2$University of Chinese Academy of Sciences \quad \\  $^3$WeChat, Tencent \quad $^4$The University of Hong Kong \\
    {\tt\small pantianlin23@mails.ucas.ac.cn} \quad {\tt\small chenyang.si@nju.edu.cn}
}

\begin{document}
\twocolumn[{
            \maketitle
            \begin{center}
                \includegraphics[width=0.8\textwidth]{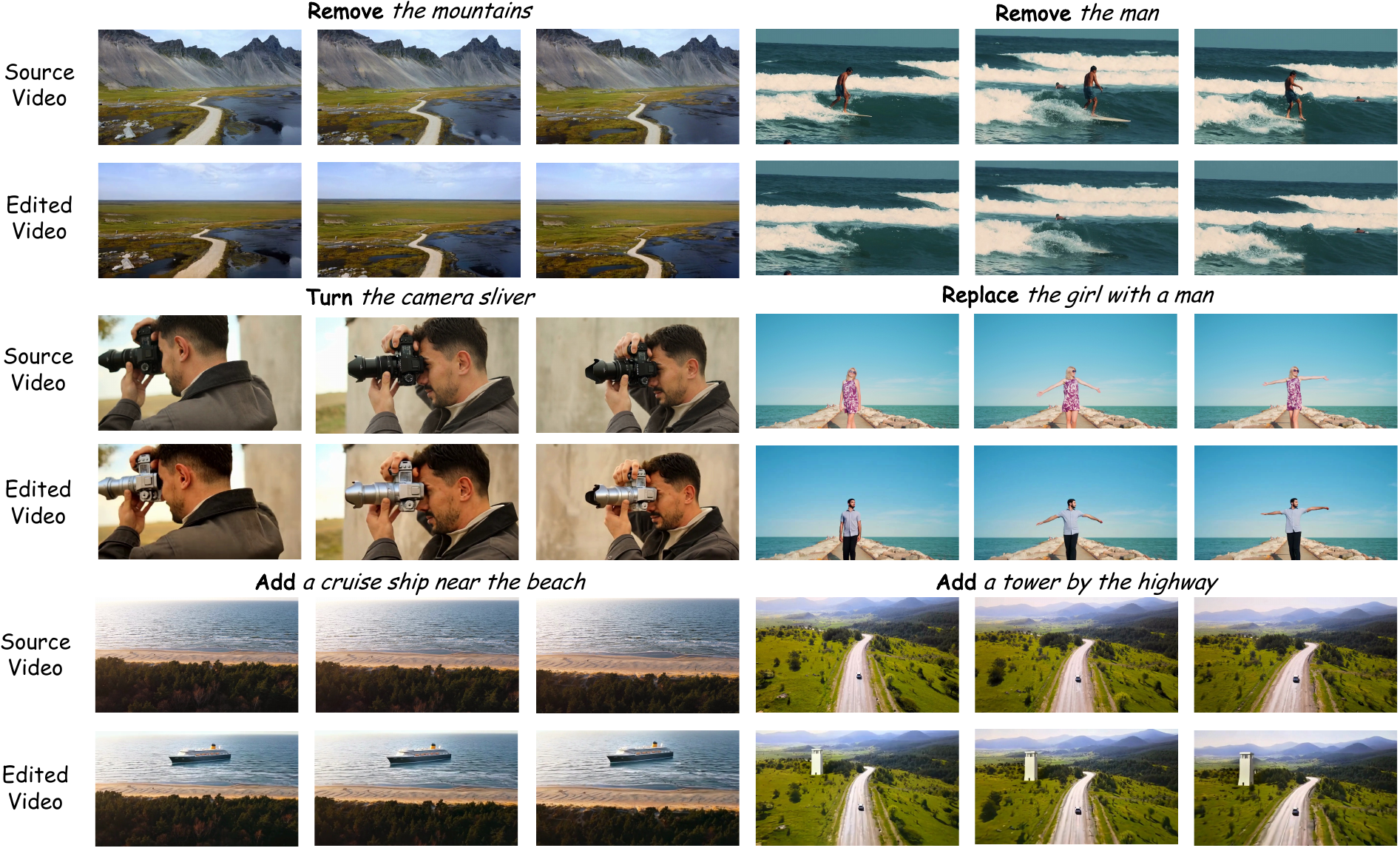}
                \captionof{figure}{
                    \textit{We propose \textbf{Sparse Control, Dense Synthesis}}, a multi-task video editing framework using \textit{sparse} user-provided keyframe edits and \textit{dense} information derived from the original video.}
                \label{fig:teaser}
            \end{center}
        }]

\renewcommand{\thefootnote}{}
\footnotetext{Work done at WeChat, Tencent Inc. $^{\dag}$Corresponding Author.}
\begin{abstract}
    Recent video editing models have achieved impressive results, but most still require large-scale paired datasets. Collecting such naturally aligned pairs at scale remains highly challenging and constitutes a critical bottleneck, especially for local video editing data. Existing workarounds transfer image editing to video through global motion control for pair-free video editing, but such designs struggle with background and temporal consistency. In this paper, we propose NOVA: Sparse Control \& Dense Synthesis, a new framework for unpaired video editing. Specifically, the sparse branch provides semantic guidance through user-edited keyframes distributed across the video, and the dense branch continuously incorporates motion and texture information from the original video to maintain high fidelity and coherence. Moreover, we introduce a degradation-simulation training strategy that enables the model to learn motion reconstruction and temporal consistency by training on artificially degraded videos, thus eliminating the need for paired data. Our extensive experiments demonstrate that NOVA outperforms existing approaches in edit fidelity, motion preservation, and temporal coherence.

    Code and models will be released at \url{https://github.com/WeChatCV/NovaEdit}.
\end{abstract}
\section{Introduction} \label{sec:intro}

\begin{figure*}[htbp]
    \centering
    \includegraphics[width=1.0\linewidth]{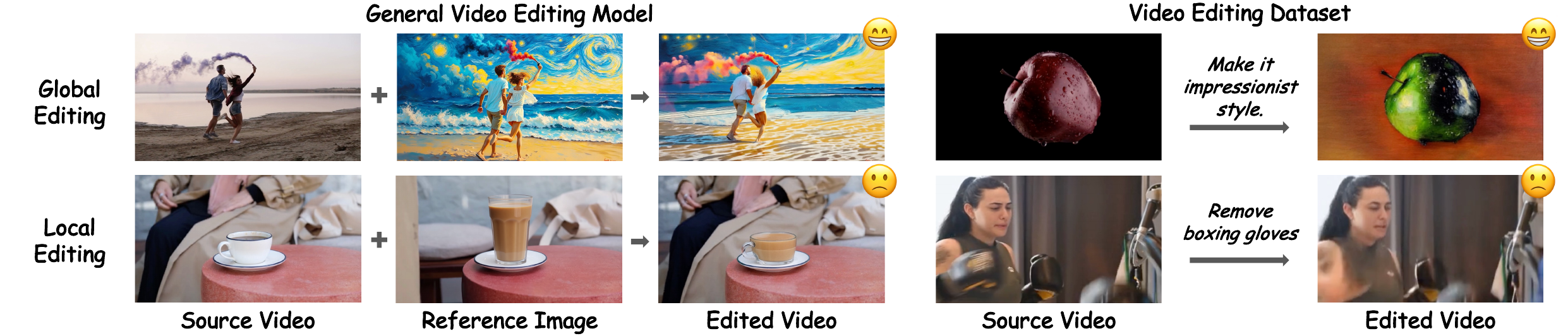}
    \caption{\textbf{Chanllenges in Local Editing}. Exisiting general video editing methods (\eg VACE~\cite{VACE}) and datasets (\eg Senorita-2M~\cite{Senorita-2M}) perform well on global editing tasks but struggle with local editing, often producing artifacts and inconsistent edits in the targeted areas.}
    \label{fig:intro}
\end{figure*}

Recent advances in diffusion models~\cite{DDPM,DDIM,DiT,LDM,U-Net,SDXL,SD3,FLUX,Sora,Veo3,HunyuanVideo,WAN} have fundamentally transformed the landscape of digital content creation. Building upon this progress, video editing~\cite{InsV2V,AnyV2V,CCEdit,VACE} is crucial for controllable and practical video content creation, yet its progress is largely constrained by data limitations. While large-scale text-video datasets for generative modeling can be collected relatively easily from open sources, constructing datasets for video editing is far more challenging, as it requires precisely aligned pairs of original and edited videos that rarely occur naturally. To address this limitation, some methods synthesize editing pairs to construct training datasets~\cite{Senorita-2M,InsViE-1M,Vivid-10M}, and then train diffusion-based video editing models on large-scale synthetic data. Other approaches~\cite{AnyV2V,I2VEdit,FlowV2V,LoRA-Edit,Senorita-2M} attempt to leverage image editing models to modify the first frame and combine the edited keyframe with motion information extracted from the source video to generate edited video sequences.

As illustrated in Figure~\ref{fig:intro}, video editing tasks can be broadly classified into two paradigms: global and local editing. We observe that existing frameworks perform well on global tasks, but face significant challenges with local video editing, which requires targeted modifications within specific spatial regions. This phenomenon can be attributed to two major bottlenecks. \textbf{1) The scarcity of high-quality paired data}: synthesizing realistic ``before'' and ``after'' video pairs for local edits is notoriously difficult, as such edits often involve complex geometric and appearance changes. Consequently, large-scale synthetic datasets usually contain visual artifacts and inconsistencies, limiting the generalization of models trained on them. \textbf{2) The fragility of first-frame-based editing}: approaches that rely solely on an edited first frame are prone to structural drift and motion misalignment, especially under significant camera or object movement. The discrepancy between the edited keyframe and the original motion dynamics often leads to degraded visual fidelity and poor temporal coherence in the resulting video. Although some techniques attempt to fine-tune a dedicated motion module for each video to improve temporal consistency, this process is labor-intensive and does not scale efficiently.

In this paper, we propose a novel framework, NOVA, that leverages sparse control and dense synthesis to learn video editing models without paired supervision. Specifically, NOVA adopts a dual-branch architecture designed to balance semantic control and fidelity. We first employ image editing models to modify keyframes within a video, where the edited keyframes serve as temporal anchors that provide stronger and more semantically aligned constraints than using a single edited frame. The Sparse Branch serves as a controllable pathway that encodes multiple edited keyframes to guide spatial and semantic transformations, imposing localized constraints on where and how edits are applied. Meanwhile, the Dense Branch encodes the unedited source video to capture dense motion and texture information. Through multi-level cross-attention, the generator queries the Dense Branch to inject fine-grained motion cues and background details, mitigating hallucination and texture drift in non-keyframe regions.

To ensure temporal coherence, NOVA integrates two complementary strategies: At \textit{inference} time, all keyframes are edited sequentially with explicit reference to the first edited frame, effectively reducing flicker and temporal discontinuities caused by independent per-frame edits. During \textit{training}, a degradation-simulation scheme is introduced to mimic real-world motion inconsistencies through interpolation and blurring of keyframes. This strategy enables the model to learn temporal restoration and texture propagation from unpaired data, thereby enhancing overall robustness and frame-to-frame consistency.

Our comprehensive experiments demonstrate that NOVA quantitatively surpasses recent approaches across all editing-specific and general quality metrics without requiring per-video fine-tuning. Furthermore, rigorous ablation studies validate the necessity of our core components, confirming Dense Synthesis Branch is crucial for background preservation via guided generative reconstruction
and that our consistency-aware inference pipeline is essential for maintaining temporal coherence.


Our contributions are summarized as follows:
\begin{itemize}
    \item We introduce and formalize the `Sparse Control, Dense Synthesis' paradigm for video editing, which first decouples dense and sparse signals for video editing, offering a new conceptual framework for the field.
    \item We design a complete unpaired learning framework, including a novel degradation-simulation training strategy and a consistency-aware inference pipeline, which together enable learning temporal coherence and motion reconstruction without any paired data.
    \item Extensive experiments demonstrate that our method outperforms state-of-the-art video editing approaches across multiple metrics.
\end{itemize}

\begin{figure*}[htbp]
    \centering
    \includegraphics[width=1.0\linewidth]{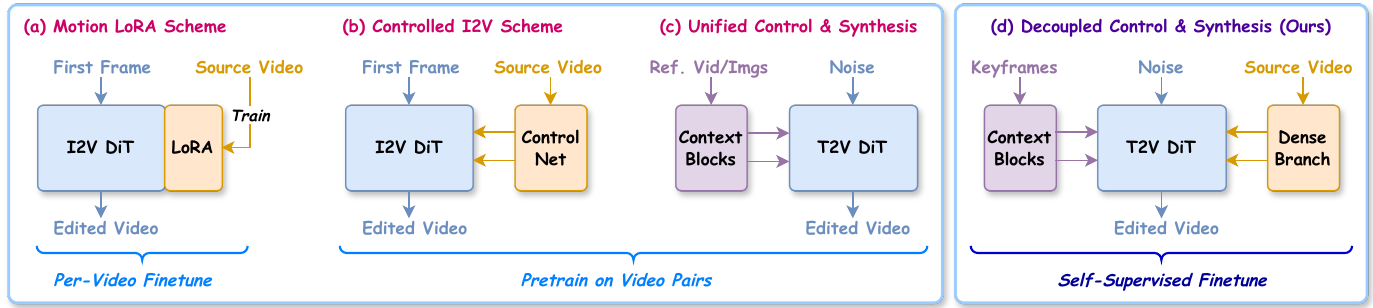}
    \caption{\textbf{The Limitations of Existing Schemes} Previous methods are often limited by either requiring costly per-video finetuning~\cite{I2VEdit,LoRA-Edit} or pre-training on large-scale paired video data~\cite{Senorita-2M,VACE}, which is difficult to acquire. Our method decouples control and synthesis signals, enabling a self-supervised framework that learns from unpaired data while maintaining high fidelity to the source video.}
    \label{fig:arch-compare}
\end{figure*}

\begin{figure}[h]
    \centering
    \includegraphics[width=1.0\linewidth]{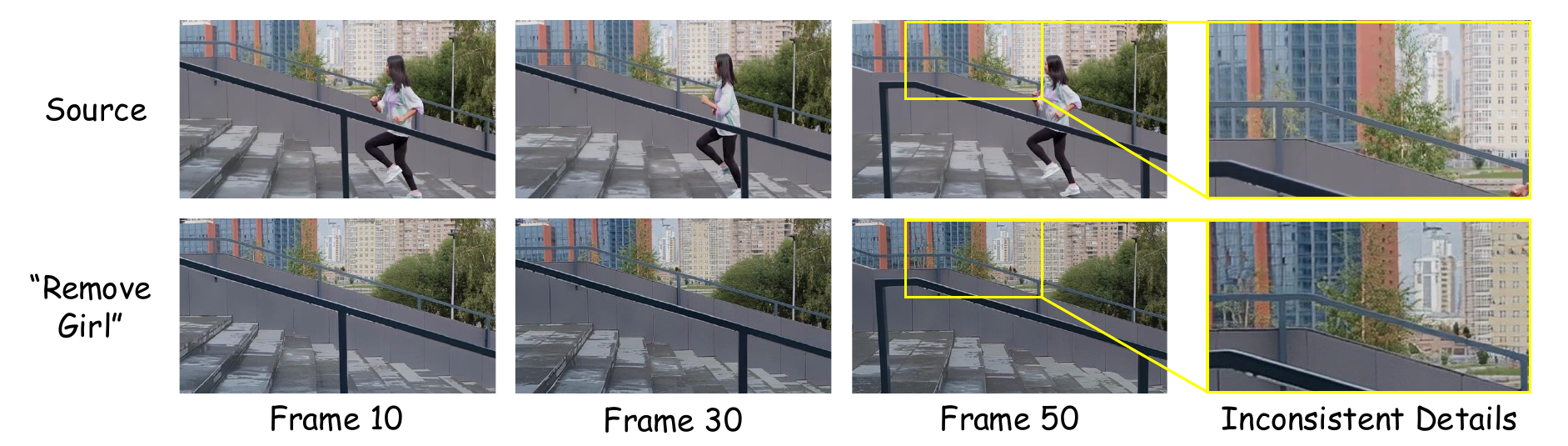}
    \caption{\textbf{Inconsistent backgrounds in naive multi-keyframe approach} We designate frames 0, 20, 40, 60, and 80 as edited keyframes (anchors). In non-keyframes reconstructed background exhibits inconsistent textures (e.g., on the building wall) and implausible motion (e.g., in the trees), as the model hallucinates content without access to the original video.}
    \label{fig:inconsistent-detail}
\end{figure}


\begin{figure*}[htbp]
    \centering
    \includegraphics[width=1.0\linewidth]{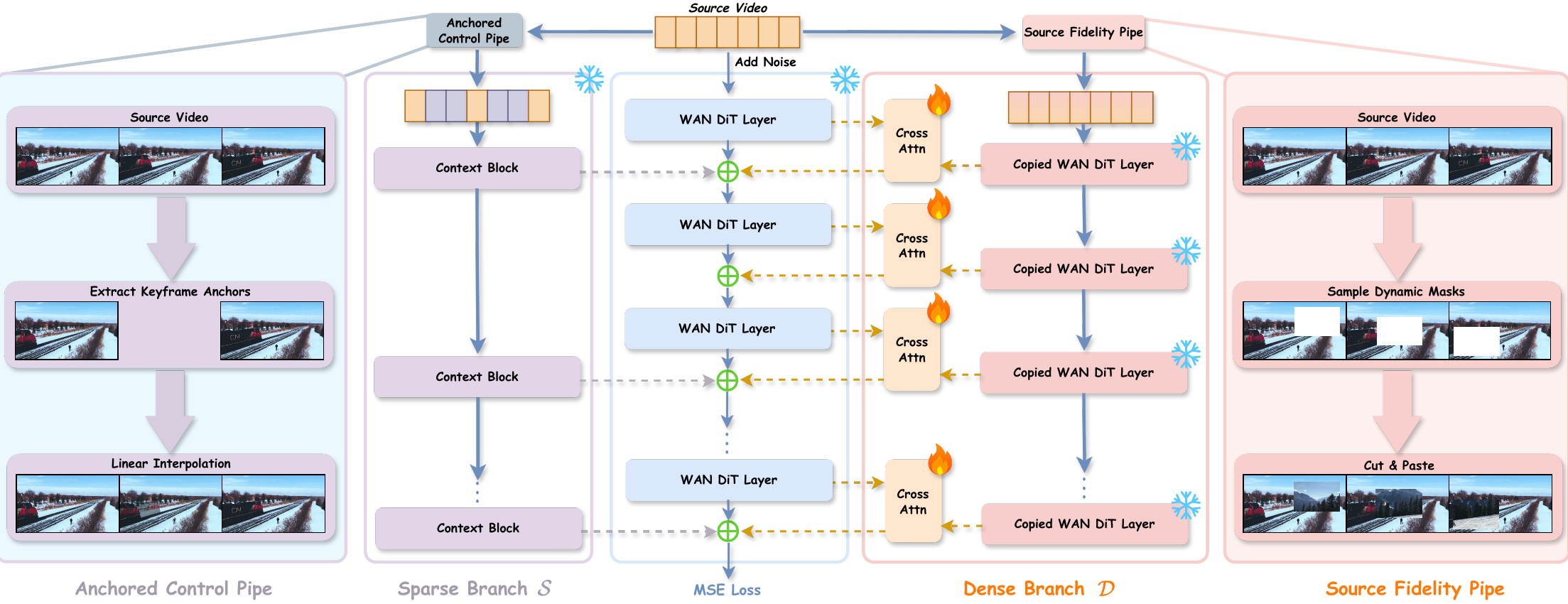}
    \caption{\textbf{Training Pipeline}. (1) \textbf{Center (Model Architecture)}: The core model learns to denoise the source video by processing conditional inputs through a Sparse Branch $\mathcal{S}$ and a Dense Branch $\mathcal{D}$, which interact via cross-attention; (2) \textbf{Left (Anchored Control Pipe)}: A degraded reference video is generated by linearly interpolating between sparsely selected keyframes, providing sparse temporal control; (3) \textbf{Right (Source Fidelity Pipe)}: A synthetic edited video is created using a cut-and-paste method to simulate realistic artifacts, serving as a dense synthesis target.}
    \label{fig:training-pipeline}
\end{figure*}
\begin{figure}[htbp]
    \centering
    \includegraphics[width=1.0\linewidth]{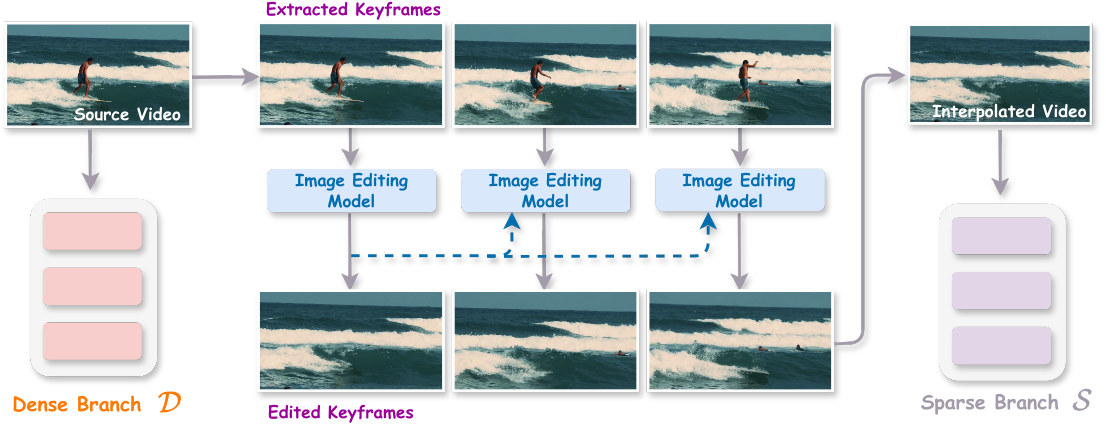}
    \caption{\textbf{Inference Pipeline}: We first edit the user-specified keyframes sequentially with reference to the first edited frame, producing temporally coherent anchors. We then construct a degraded reference video by interpolating between these keyframes, and feed it into the sparse branch $\mathcal{S}$. The original unedited video is passed through the dense branch $\mathcal{D}$.}
    \label{fig:inference-pipeline}
\end{figure}
\begin{figure*}[htbp]
    \centering
    \includegraphics[width=0.95\linewidth]{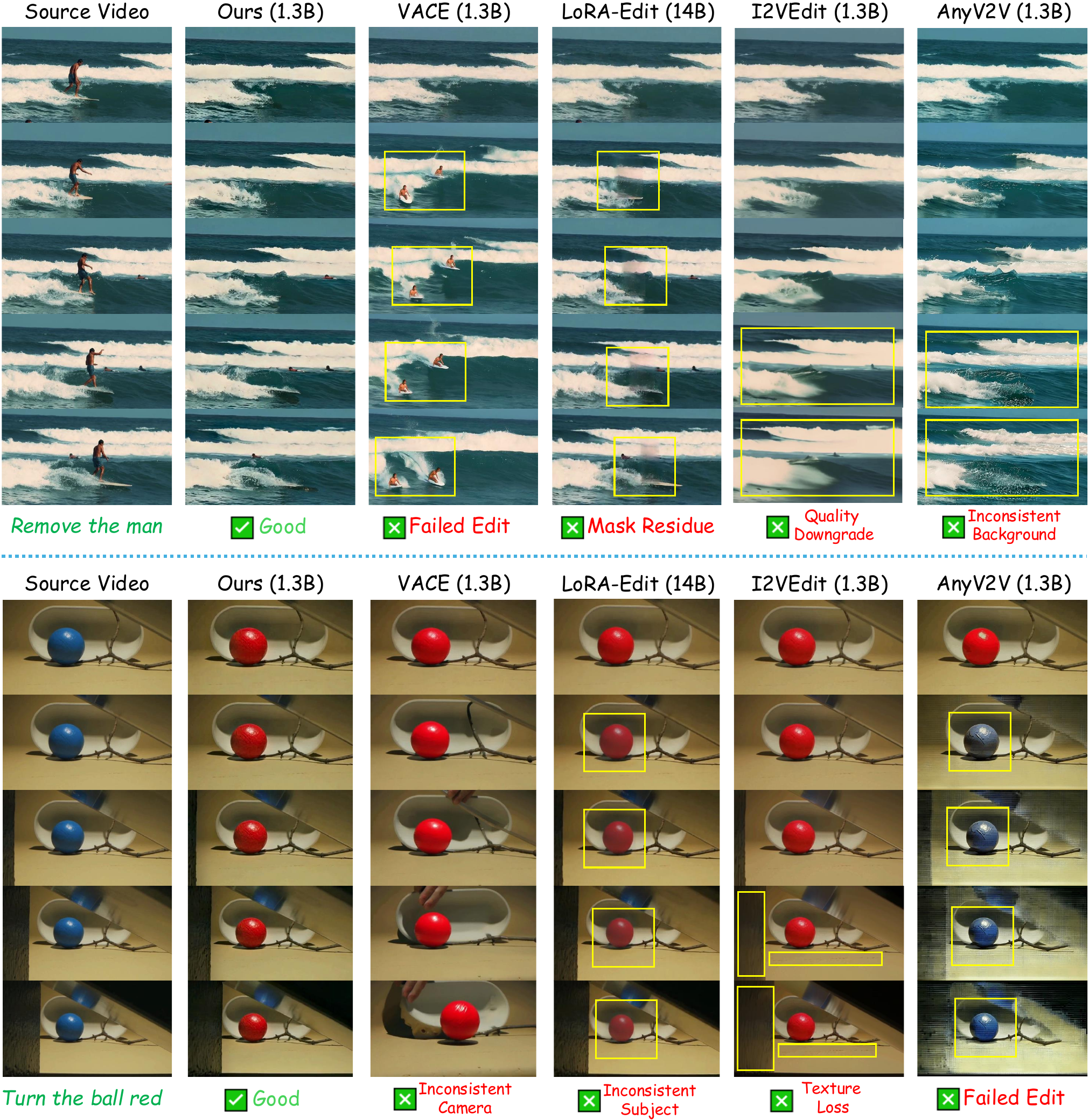}
    \caption{Qualitative Comparisons with VACE~\cite{VACE}, LoRA-Edit~\cite{LoRA-Edit}, I2VEdit~\cite{I2VEdit} and AnyV2V~\cite{AnyV2V}.}
    \label{fig:compare}
\end{figure*}

\section{Related Work}
\label{sec:relatedwors}
\subsection{Video Editing with Diffusion Models}
The success of text-guided image editing~\cite{InsPix2Pix,Prompt2Prompt,VQGAN-CLIP,EditGAN,FLUX-Kontext,Qwen-Image,Nano-Banana} has inspired significant efforts in the video domain. A primary challenge has been maintaining temporal consistency across frames. To address this, various strategies have been proposed, such as enforcing coherence in diffusion features~\cite{TokenFlow}, leveraging optical flow in attention mechanisms~\cite{FLATTEN}, merging self-attention tokens~\cite{VidToMe}, or learning transferable motion priors~\cite{AnimateDiff}. Concurrently, zero-shot methods have gained traction, enabling video editing without task-specific training by propagating edits from an initial frame or using pre-trained image models~\cite{RerenderAVideo,Pix2Video,Vid2Vid-Zero}. To enhance user control, subsequent works introduced more precise manipulation techniques, including cross-attention control~\cite{Video-P2P}, guidance via explicit control signals~\cite{Control-A-Video} and drag-style editing~\cite{DragVideo}. Research has also branched into specialized applications like video inpainting, with dedicated frameworks for filling masked regions consistently~\cite{ProPainter,AVID,CoCoCo}. More recently, the field is trending towards unified, multi-task models capable of handling diverse video understanding and generation tasks within a single framework~\cite{O-DisCo-Edit,OmniVideo,UniVid}, while others focus on high-speed synthesis~\cite{Fairy}.

\subsection{Data Synthesis for Video Editing}
To address the challenges of data scarcity in video editing, several approaches have been proposed to synthesize training data. DreamVE~\cite{DreamVE}, MiniMax-Remover~\cite{MiniMax-Remover}, and VACE~\cite{VACE} utilize segmentation models like Grounding DINO~\cite{GroundingDINO} and SAM2~\cite{SAM2} to create synthetic video pairs by combining cut-and-paste techniques. ReCamMaster~\cite{ReCamMaster} and ROSE~\cite{ROSE} leverage game engines to generate diverse video datasets with controllable attributes. Moreover, Senorita-2M~\cite{Senorita-2M} introduces several video experts to generate a large-scale video editing dataset for different tasks.

\subsection{Frame-Guided Video Editing}
Another way to mitigate the data scarcity issue is to use frame-guided editing methods~\cite{AnyV2V,StableV2V,FlowV2V,LoRA-Edit,I2VEdit,GenProp,VACE}. These methods typically use an image-to-video (I2V) model conditioned on the edited first frame to propagate the edit to the entire video. To maintain consistency with the original video, AnyV2V~\cite{AnyV2V} reconstructs motion via DDIM inversion and temporal feature injection. FlowV2V~\cite{FlowV2V} and VACE~\cite{VACE} incorporate motion cues like optical flow and depth maps to guide the editing process. I2VEdit~\cite{I2VEdit} and LoRA-Edit~\cite{LoRA-Edit} fine-tune a unique motion adapter (LoRA~\cite{LoRA}) for each video to achieve better consistency. While these methods have shown promising results, they often struggle with addition and removal tasks or require significant computational resources for LoRA fine-tuning.
\section{Methodology}
\subsection{The Limitations of Existing Paradigms}
Current video editing paradigms are constrained by their underlying architectural designs, as shown in Figure~\ref{fig:arch-compare}. First-frame-guided architectures in Figure~\ref{fig:arch-compare} (a) and (b) treat editing as a propagation problem. They rely on a single edited frame as a temporal anchor and task the model with extending this edit across the entire sequence. Whether through per-video motion adapters or explicit control modules, these architectures fundamentally entangle appearance generation with motion preservation. This tight coupling makes them fragile; any small discrepancy between the initial edit and the source video's dynamics can compound over time, leading to structural drift and temporal incoherence.

Alternatively, unified generative architectures in Figure~\ref{fig:arch-compare} (c) merge control signals and synthesis into a single pathway. While conceptually simple, this approach places an enormous burden on the model to disentangle what to change from what to preserve. As demonstrated in Figure~\ref{fig:inconsistent-detail}, even a multi-keyframe approach fails if the architecture lacks a mechanism to preserve source fidelity. This motivates our shift in architecture towards \textbf{decoupling} control from synthesis. We propose the Sparse Control, Dense Synthesis paradigm (Figure~\ref{fig:arch-compare} (d)), which leverages sparse, edited keyframes for semantic guidance while using the dense, original video to preserve motion and texture integrity. This approach resolves the core bottlenecks of previous methods, enabling high-quality editing without reliance on paired data.

\subsection{Decoupling Control and Synthesis}

As illustrated in Figure~\ref{fig:training-pipeline}, we design our architecture to integrate multi-keyframe sparse control with dense synthesis signals. It consists of three main components: a main denoising branch, a sparse control branch, and a dense synthesis branch.
The sparse branch, composed of several WAN DiT layers, injects edited keyframe information into the main branch to guide the editing process.
The dense synthesis branch, structurally identical to the main branch, transfers motion and texture cues from the source video, facilitating the reconstruction of motion and non-edited regions in the main branch.

To prevent direct fusion of dense-branch features from interfering with the editing process, we additionally introduce a series of trainable cross-attention modules between corresponding layers of the dense and main branches. During the forward pass, the main branch generates queries, while the dense branch provides keys and values. The resulting cross-attention output is fused back into the main branch. This interaction can be formulated as follows:
\begin{align}
    \boldsymbol{z}^{(l)}_{\text{m}} \leftarrow
    \boldsymbol{z}^{(l)}_{\text{m}}
    + \underbrace{\mathcal{S}^{(l)}(\boldsymbol{z}^{(l)}_{\text{m}}, \boldsymbol{r})}_{\substack{\text{Sparse Control}                   \\ \text{  from edited keyframes}}}
    + \underbrace{\mathcal{D}^{(l)}(\boldsymbol{z}^{(l)}_{\text{m}}, \boldsymbol{z}^{(l)}_{\text{d}})}_{\substack{\text{Dense Synthesis} \\ \text{ from source video}}},
\end{align}
where $\boldsymbol{z}^{(l)}_{\text{m}}$ and $\boldsymbol{z}^{(l)}_{\text{d}}$ denote the latent features at layer $l$ of the main and the copied DiTs, and $\mathcal{S}^{(l)}$ and $\mathcal{D}^{(l)}$ represent the output of the sparse VACE~\cite{VACE} block and the cross-attention module at layer $l$, respectively. $\boldsymbol{r}$ is the reference condition constructed from the edited keyframes, as described in Section~\ref{sec:train-pipe} (during training) and~\ref{sec:infer-pipe} (during inference).

\subsection{Learning Coherence without Paired Data}
\label{sec:train-pipe}
To enable training without paired data, we synthesize self-supervised signals that simulate real editing behaviors. Specifically, each training sample is processed through two complementary pipelines: the Anchored Control Pipeline, which constructs degraded edited keyframe sequences for sparse supervision, and the Source Fidelity Pipeline, which generates synthetic source videos for dense supervision.

\noindent\textbf{Anchored Control Pipeline} samples and degrades a sparse set of keyframes from the target video through a series of synthetic degradations and interpolation. This process emulates the low-quality or inconsistent sequences that can result from manual or automated keyframe editing, and these degraded keyframes serve as inputs to the sparse branch to guide the editing process.
Formally, given the target video $\mathcal{X} = \{\boldsymbol{x}_t\}_{t=0}^T$, we first select a sparse set of keyframe indices $\mathcal{K} = \{k_0, k_1, \ldots, k_N\} \subseteq \{0, 1, \ldots, T\}$ with $k_0 = 0$ and $k_N = T$.
To simulate a diverse range of local editing artifacts, including appearance mismatches and geometric misalignments (e.g., warping, zooming, and jitter), a random subset of keyframes undergoes a stochastic degradation process. This is applied to localized regions specified by binary masks $\boldsymbol{b}_{k_i} \in \{0,1\}^{H\times W\times1}$.
Each corrupted keyframe $\hat{\boldsymbol{x}}_{k_i}$ is generated as:
\begin{equation}
    \hat{\boldsymbol{x}}_{k_i}
    = (\boldsymbol{1} - \boldsymbol{b}_{k_i}) \odot \boldsymbol{x}_{k_i}
    + \boldsymbol{b}_{k_i} \odot \mathcal{D}_{\text{aug}}(\boldsymbol{x}_{k_i}),
    \quad \forall\, k_i \in \mathcal{K} \backslash \{0\},
\end{equation}
where $\mathcal{D}_{\text{aug}}$ is a stochastic degradation operator that applies a composition of common image corruptions, such as Gaussian blurring and random affine transformations, to the input frame, and $\odot$ represents element-wise multiplication.
A motion-degraded video $\hat{\mathcal{X}} = \{\hat{\boldsymbol{x}}_t\}_{t=0}^T$ is then reconstructed by linearly interpolating between adjacent degraded keyframes:
\begin{equation}
    \hat{\boldsymbol{x}}_t = \begin{cases}
        \hat{\boldsymbol{x}}_{k_n},                                                            & t \in \mathcal{K},   \\
        (1 - \alpha_t) \hat{\boldsymbol{x}}_{k_{n-1}} + \alpha_t \hat{\boldsymbol{x}}_{k_{n}}, & k_{n - 1} < t < k_n,
    \end{cases}\label{eq:interpolation}
\end{equation}
where $\alpha_t = \frac{t - k_{n-1}}{k_n - k_{n-1}}$.
This interpolated sequence, containing both motion degradation (from missing dynamics between keyframes) and appearance inconsistencies (from simulated editing noise), is used as the edited reference $\boldsymbol{r}$ for the sparse branch $\mathcal{S}$.

\noindent\textbf{Source Fidelity Pipeline} applies a random Cut-and-Paste strategy to the target video, producing a spatially misaligned pseudo source video that mimics the unedited input during inference. This pseudo source video is fed into the dense branch to provide motion and texture references.
Specifically, we generate a synthetic \textbf{source} video by pasting randomly sampled content onto the target frames through a moving binary mask that undergoes random translation and rotation.
Formally, let a target video $\mathcal{X} = \{\boldsymbol{x}_t\}_{t=0}^T$ and another randomly sampled video $\mathcal{Y} = \{\boldsymbol{y}_t\}_{t=0}^T$, both of length $T\!+\!1$ with $\boldsymbol{x}_t, \boldsymbol{y}_t \in \mathbb{R}^{H \times W \times C}$.
We define a temporally coherent binary mask sequence $\mathcal{M} = \{\boldsymbol{m}_t\}_{t=0}^T$, where each $\boldsymbol{m}_t \in \{0,1\}^{H \times W \times 1}$ is obtained by applying random affine transformations (translation, rotation, and scaling) to a base shape (e.g., ellipse or polygon).
The pseudo source video $\tilde{\mathcal{X}} = \{\tilde{\boldsymbol{x}}_t\}_{t=0}^T$ is then constructed as
\begin{equation}
    \tilde{\boldsymbol{x}}_{t}
    = \boldsymbol{m}_{t} \odot \boldsymbol{y}_{t}
    + \bigl(\boldsymbol{1} - \boldsymbol{m}_{t}\bigr) \odot \boldsymbol{x}_{t},
    \qquad \forall\, t \in \{0,\dots,T\},
\end{equation}
where $\odot$ denotes element-wise multiplication, $\boldsymbol{1}\in\mathbb{R}^{H\times W\times1}$ is an all-ones tensor, and $\boldsymbol{m}_t$ is broadcast along the channel dimension to match $\boldsymbol{x}_t, \boldsymbol{y}_t$.
This pseudo source video serves as the input to the dense branch $\mathcal{D}$.

Together with the Source Fidelity Pipeline, this design enables two complementary capabilities:
(a) recovering natural motion and texture via the dense branch $\mathcal{D}$, and
(b) maintaining temporal coherence across keyframes via the sparse branch $\mathcal{S}$.
The overall training objective adopts a standard denoising loss:
\begin{equation}
    \mathcal{L}
    = \mathbb{E}_{\mathcal{X}, \epsilon, t}
    \!\left[
        \| \epsilon - \epsilon_\theta(\boldsymbol{z}_t(\mathcal{X}), t, \tilde{\mathcal{X}}, \hat{\mathcal{X}}) \|_2^2
        \right],
\end{equation}
where $\tilde{\mathcal{X}}$ and $\hat{\mathcal{X}}$ denote the pseudo source and degraded edited videos, and $\boldsymbol{z}_t$ is the noisy latent at timestep $t$ derived from the original video $\mathcal{X}$.

\subsection{Inference with Multi-Keyframe Guidance}
\label{sec:infer-pipe}
\noindent\textbf{Consistency-Aware Keyframe Editing.}
To ensure visual consistency across independently edited frames, we adopt a consistency-aware keyframe editing scheme. Specifically, we employ the FLUX.1 Kontext Inpainting model~\cite{FLUX}, which supports conditioning on a reference image to preserve appearance across edits.
For the first keyframe ($t=0$), we perform standard text-guided image-to-image editing based on the user prompt.
For subsequent keyframes, each edit is conditioned on the result of the first edited keyframe to maintain stylistic coherence:
\begin{equation}
    \boldsymbol{x}^{\mathrm{edit}}_{k_i}
    = \text{FLUX}\!\left(\boldsymbol{x}_{k_i}, \boldsymbol{x}^{\mathrm{edit}}_{k_0}, \boldsymbol{m}_{k_i}, \mathcal{P}\right),
    \quad i = 1, \ldots, N,
\end{equation}
where $\boldsymbol{m}_{k_i}$ denotes the optional user-provided mask and $\mathcal{P}$ is the text prompt.
By anchoring all keyframe edits to the first edited frame, we greatly reduce appearance drift and flickering in the final video.

\noindent\textbf{Constructing Reference and Source Inputs.}
After obtaining the edited keyframes $\{\boldsymbol{x}^{\mathrm{edit}}_{k_i}\}$, we generate an interpolated reference video $\hat{\mathcal{X}}$ by linearly interpolating between adjacent keyframes, following Eq.~\ref{eq:interpolation}.
This temporally completed sequence serves as the reference condition $\boldsymbol{r}$ for the sparse branch $\mathcal{S}$.
Meanwhile, the original unedited video $\mathcal{X}$ is processed by the dense DiT branch $\mathcal{D}$ to supply motion and texture information.
This dual-input inference design mirrors the training setup, ensuring that the model faithfully preserves source dynamics while applying user-specified edits at key moments.

\section{Experiments}
\subsection{Implementation Details}
\label{sec:imple-detail}
Our proposed framework is built upon WAN 2.1 VACE 1.3B~\cite{VACE} architecture. To efficiently integrate the new Dense Synthesis capability, we trained only the newly introduced cross-attention modules connecting the main DiT and the dense DiT branch, keeping the weights of the base WAN 2.1 VACE model frozen throughout training. We trained our model on a dataset consisting of $5,000$ high-quality video clips sourced from Pexels~\cite{Pexels}. The training process utilized total of 320GB aggregate GPU memory. We employed the AdamW optimizer and trained for approximately $8,000$ steps with a fixed learning rate of $1 \times 10^{-4}$. All videos were processed at a resolution of $832 \times 480$ pixels and frame length $81$. During inference, we use FLUX.1 Kontext Dev~\cite{FLUX-Kontext} to generate edited keyframes, as detailed in Section~\ref{sec:infer-pipe}. We utilize keyframes at indices $\{0, 10, \ldots, 80\}$ for a video length of 81.

\begin{table*}[htbp]
    \centering
    \begin{tabular}{lcccccc cc}
        \toprule
        \multirow{2}{*}{\textbf{Method}}                      &
        \multirow{2}{*}{\textbf{Param}}                       &
        \multirow{2}{*}{\textbf{\makecell{Per Vid.                                                                                                                                                                                          \\ Finetune}}} &
        \multicolumn{1}{c}{\textbf{Human Eval.}}              &
        \multicolumn{3}{c}{\textbf{Editing-Specific Metrics}} &
        \multicolumn{2}{c}{\textbf{VBench}}                                                                                                                                                                                                 \\
        \cmidrule(lr){4-4} \cmidrule(lr){5-7} \cmidrule(lr){8-9}
                                                              &      &           & \textbf{SR} $\uparrow$ & \textbf{TC} $\uparrow$ & \textbf{FC} $\uparrow$ & \textbf{BG-SSIM} $\uparrow$ & \textbf{MS} $\uparrow$ & \textbf{BC} $\uparrow$ \\
        \midrule
        AnyV2V~\cite{AnyV2V}                                  & 1.3B & \ding{55} & 0.75                   & 0.918                  & 0.840                  & 0.858                       & 0.973                  & 0.939                  \\
        \midrule
        I2VEdit~\cite{I2VEdit}                                & 1.3B & \ding{51} & 0.83                   & 0.931                  & 0.846                  & 0.900                       & \underline{0.991}      & {0.941}                \\
        \midrule
        LoRA-Edit~\cite{LoRA-Edit}                            & 14B  & \ding{51} & 0.80                   & 0.923                  & \underline{0.880}      & 0.901                       & 0.984                  & 0.929                  \\
        LoRA-Edit (\textit{Multi-keyframe})                   & 14B  & \ding{51} & 0.86                   & \underline{0.933}      & 0.869                  & 0.909                       & 0.986                  & 0.933                  \\
        \midrule
        VACE~\cite{VACE}                                      & 1.3B & \ding{55} & 0.36                   & 0.926                  & 0.817                  & 0.807                       & \underline{0.991}      & 0.923                  \\
        VACE (\textit{Multi-keyframe})                        & 1.3B & \ding{55} & \underline{0.90}       & 0.928                  & 0.840                  & 0.913                       & 0.989                  & 0.940                  \\
        \midrule
        Senorita-2M~\cite{Senorita-2M}                        & 5B   & \ding{55} & 0.86                   & 0.919                  & 0.853                  & \textbf{0.921}              & 0.989                  & \textbf{0.953}         \\
        \midrule
        \textbf{Ours}                                         & 1.3B & \ding{55} & \textbf{0.93}          & \textbf{0.935}         & \textbf{0.882}         & \underline{0.917}           & \textbf{0.993}         & \underline{0.946}      \\
        \bottomrule
    \end{tabular}
    \caption{\textbf{Quantitative comparison with state-of-the-art methods}: We compare our method with recent frame-guided video editing approaches on multiple metrics. Our method outperforms all baselines in most metrics while requiring no per-video finetuning.}
    \label{tab:quantitative_comparison}
\end{table*}

\subsection{Main Results}
We compare our method with recent frame-guided video editing approaches, including AnyV2V~\cite{AnyV2V}, I2VEdit~\cite{I2VEdit}, LoRA-Edit~\cite{LoRA-Edit}, VACE~\cite{VACE} and Senorita-2M~\cite{Senorita-2M}. For test videos, we follow I2VEdit~\cite{I2VEdit} and Render-A-Video~\cite{RerenderAVideo} to collect videos from Pexels~\cite{Pexels}. For fairness, we also tested the performance of LoRA-Edit and VACE when provided with additional frame guidance. We ensured that the content of edited keyframes were identical for all models.

\noindent\textbf{Qualitative Comparison} The visual comparison presented in Figure~\ref{fig:compare} demonstrates that our method offers distinct advantages over the baselines, particularly concerning background consistency and overall visual fidelity.


\noindent\textbf{Quantitative Comparison}
Following I2VEdit~\cite{I2VEdit} and LoRA-Edit~\cite{LoRA-Edit}, we first apply Temporal Consistency (\textbf{TC}) to quantify the semantic alignment throughout the generated sequence by measuring the CLIP~\cite{CLIP} embedding similarity between the generated frames and the edited first frame. Frame Consistency (\textbf{FC}) is also introduced as the per-frame CLIP embedding similarity between the generated output frames and the original input frames, measuring fidelity to the source content. In addition, we introduce Success Rate (\textbf{SR}), a critical metric from our user study, which is determined by asking participants whether the model successfully propagated the first frame's editing result to all subsequent frames in the sequence. As FC is ill-suited for object removal or addition tasks, we also introduce Background SSIM (\textbf{BG-SSIM}) to compute the structural similarity on the unedited background regions, which are determined by masks from SAM2~\cite{SAM2}. Finally, we incorporate two metrics from VBench~\cite{VBench}: Motion Smoothness (\textbf{MS}) and Background Consistency (\textbf{BC}).

\subsection{Ablation Studies}

\begin{figure}[h]
    \centering
    \includegraphics[width=1.0\linewidth]{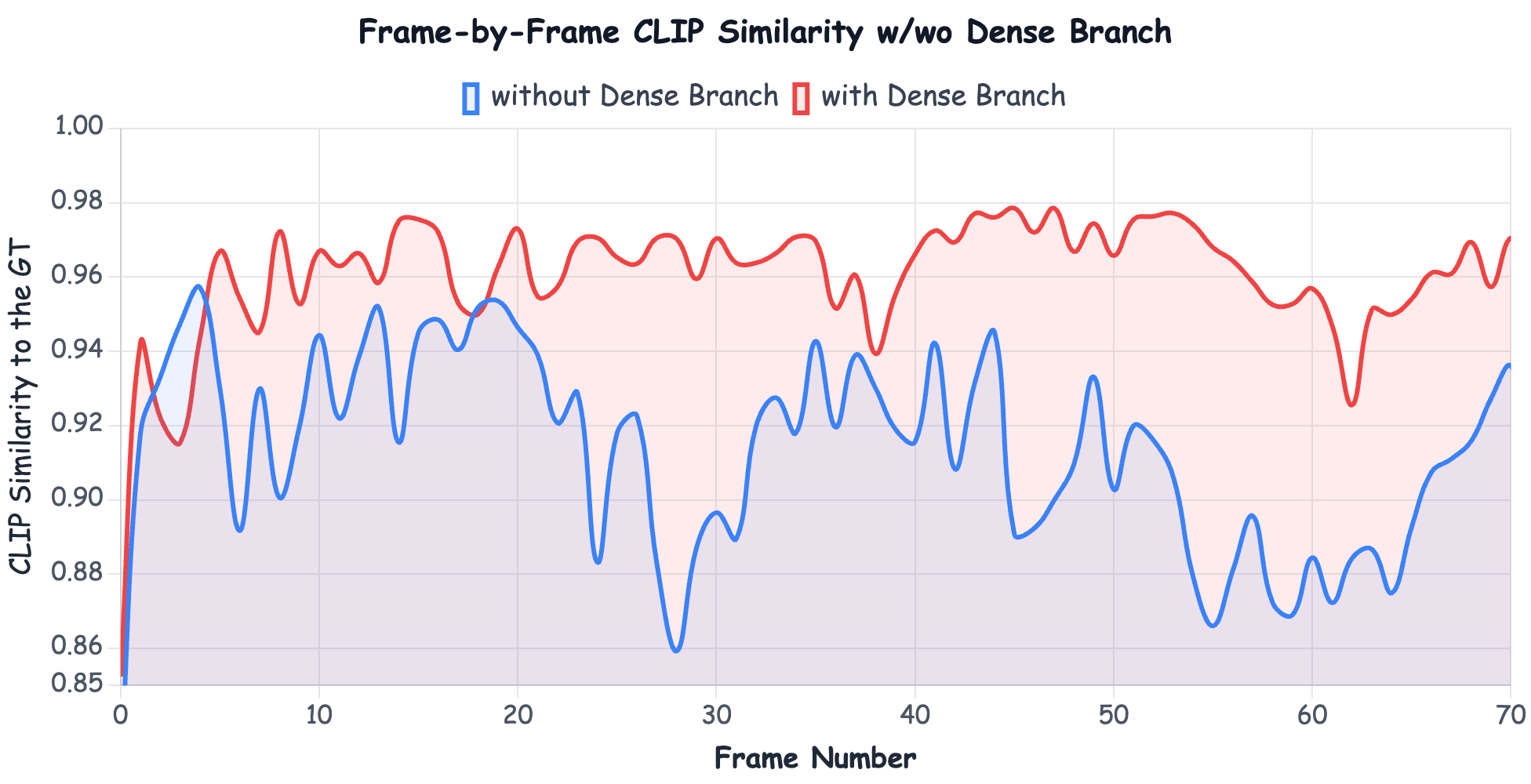}
    \vspace{-1.5em}
    \caption{\textbf{Ablation of the Dense Branch}: The inclusion of the Dense Branch enables the model to capture background details within the input video, thereby significantly enhancing the per-frame CLIP Similarity with the ground truth.}
    \vspace{-1em}
    \label{fig:aba-dense-branch}
\end{figure}

\begin{figure}[h]
    \centering
    \includegraphics[width=1.0\linewidth]{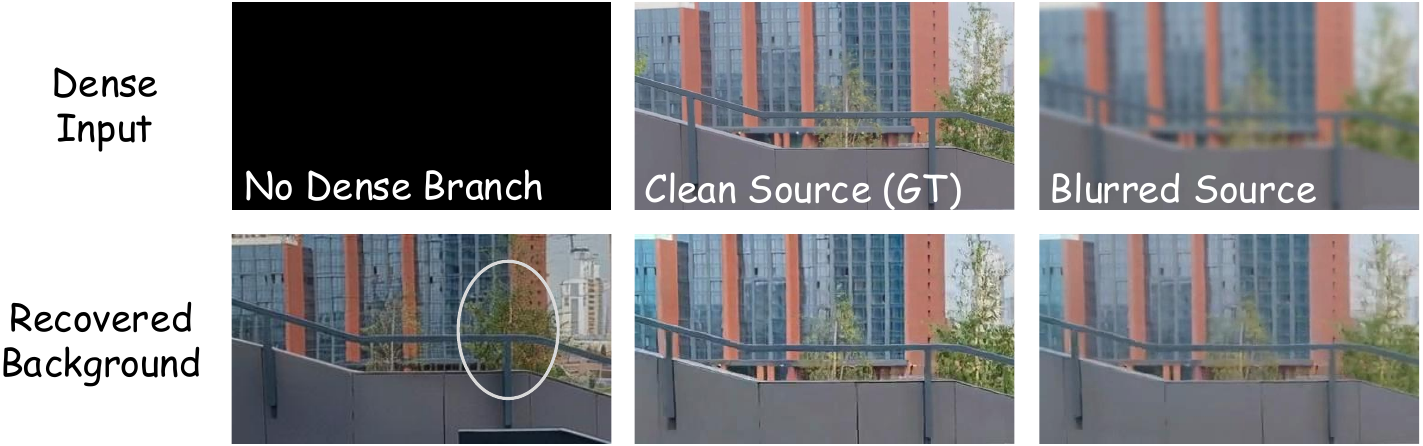}
    \vspace{-1em}
    \caption{
        \textbf{Ablation on the Role and Robustness of the Dense Branch.}
        \textbf{(Left)} Deactivating the dense branch entirely results in hallucinated details in the background.
        \textbf{(Middle)} Providing the clean source video enables a near-perfect reconstruction that preserves high-frequency textures.
        \textbf{(Right)} The model recovers a background that is visibly sharper than the blurred input.
    }
    \vspace{-1em}
    \label{fig:ablation_dense_branch}
\end{figure}

\noindent\textbf{The Effect of Dense Branch}
To quantitatively validate the necessity of our dense branch, we construct an object addition/removal dataset by leveraging SAM2~\cite{SAM2} to generate high-fidelity object masks, creating ground truth video pairs via a cut-and-paste methodology. We then compare our full model against an ablated baseline where the dense branch is removed. We evaluate performance by computing the average frame-wise CLIP Similarity between the generated video and the ground truth. As shown in Figure~\ref{fig:aba-dense-branch}, introducing the Dense Branch can improve background consistency in the edited video. Furthermore, we perform a qualitative ablation to probe its mechanism and robustness by varying the quality of its video input at inference time. The results, detailed in Figure~\ref{fig:ablation_dense_branch}, confirm that the dense branch is essential for preserving background consistency. More importantly, the experiment shows that the model can robustly reconstruct details even from a degraded source, confirming that its function extends beyond simple texture copying to a more sophisticated, guided synthesis.

\begin{figure}[h]
    \centering
    \includegraphics[width=1.0\linewidth]{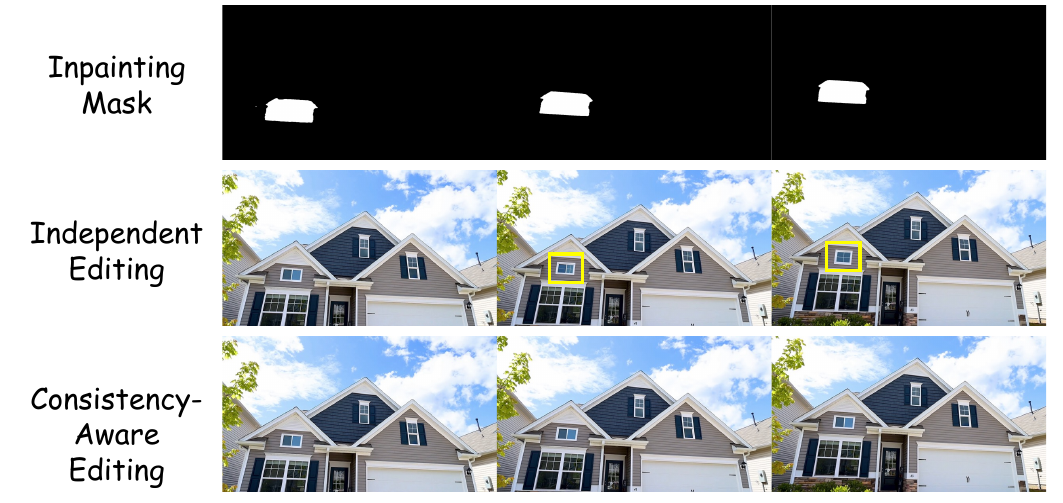}
    \vspace{-1.5em}
    \caption{{Ablation of Consistency-Aware Keyframe Editing}}
    \vspace{-1em}
    \label{fig:aba-infer}
\end{figure}

\noindent\textbf{The Effect of Consistency-Aware Keyframe Editing}
To validate our inference pipeline, we test a baseline that edits each keyframe independently without referencing the first edited frame. As shown in Figure~\ref{fig:aba-infer}, when editing independently, the added window exhibits noticeable inconsistencies in styling; however, using the edited first frame as a reference significantly improves consistency.


\begin{table}[h]
    \centering
    \begin{tabular}{lccc}
        \toprule
        \textbf{Model}       & \textbf{TC} $\uparrow$ & \textbf{FC} $\uparrow$ & \textbf{BG-SSIM} $\uparrow$ \\
        \midrule
        FLUX.1-Kontext-dev   & \textbf{0.92}          & 0.85                   & 0.88                        \\
        Qwen-Image-Edit-2509 & 0.88                   & \textbf{0.87}          & \textbf{0.89}               \\
        \bottomrule
    \end{tabular}
    \vspace{-0.5em}
    \caption{{Ablation Study of Keyframe Editing Models.}}
    \label{tab:flux-qwen-ablation}
\end{table}

\noindent\textbf{The Sensitivity of Keyframe Editing Model}
We evaluate the framework's sensitivity to the choice of the keyframe editing model by performing an ablation where FLUX.1 Kontext is replaced by Qwen-Image-Edit. The results presented in Table~\ref{tab:flux-qwen-ablation} indicate that our overall pipeline is not strictly coupled to a single model, underscoring the generality of our paradigm.

\begin{figure}[h]
    \centering
    \includegraphics[width=1.0\linewidth]{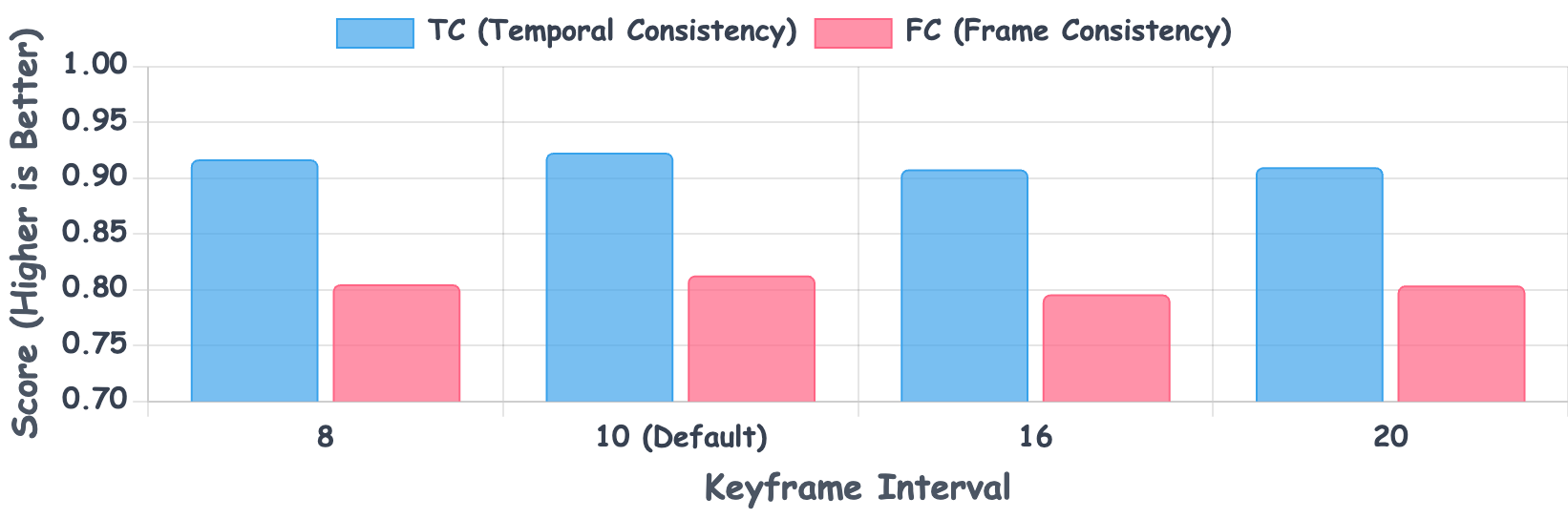}
    \vspace{-2em}
    \caption{\textbf{Ablation study on the keyframe interval used during inference}: The model was trained with a fixed interval of $10$. We evaluate its robustness by testing with different intervals.}
    \vspace{-1em}
    \label{fig:ablation_interval}
\end{figure}

\noindent\textbf{The Sensitivity of Keyframe Interval at Inference}
Our model is trained using a fixed keyframe interval of $10$ frames (See Section~\ref{sec:imple-detail}). To assess the model's robustness and its ability to generalize to different levels of guidance sparsity, we conduct an ablation study on the keyframe interval used during inference. We vary the interval between keyframes, testing intervals of $8$, $16$, and $20$ (divisors of $80$), in addition to the default training interval of $10$. The results, presented in Figure~\ref{fig:ablation_interval}, demonstrate the model's robustness to variations in the frequency of keyframe guidance. It exhibits that our model is not overfitted to a specific keyframe frequency and can flexibly handle user-provided keyframes at different levels of sparsity.
\section{Conclusion}
In this paper, we introduced Sparse Control, Dense Synthesis, a novel paradigm for video editing that effectively balances user-guided semantic changes with the preservation of original video dynamics. Our dual-branch architecture, featuring a Sparse Branch for multi-keyframe guidance and a Dense Branch for source detail injection, successfully addresses the common problem of motion and texture hallucination in non-edited regions. We believe this paradigm offers a promising direction for future research in high-quality and efficient video editing.

\noindent\textbf{Limitations}: The performance of our method is influenced by the quality of the edited keyframe anchors. While our context-aware pipeline improves temporal consistency, the single-pass generation of high-fidelity keyframes can be challenging for current image editing models, which may require user iteration to achieve optimal results.
\section{Acknowledgement}
This study is partially supported by Beijing Natural Science Foundation (QY25188), the Fundamental Research Funds for the Central Universities (KG2025XX) and Jiangsu Science and Technology Major Project (BG2025035). This research is also supported by cash and in-kind funding from Nanjing K\&A Center of Cultivation and industry partner(s).
{
    \small
    \bibliographystyle{ieeenat_fullname}
    \bibliography{main}
}

\clearpage
\appendix
\setcounter{page}{1}
\maketitlesupplementary

\section{Details of Experiments on Naive Multi-Keyframe Guidance}
\label{sec:motivation_exp}

To explore the feasibility of multi-keyframe guidance for temporally consistent video editing, we conduct preliminary experiments using the reference-to-video diffusion model WAN VACE. VACE accepts two auxiliary inputs: (1) a \textit{reference video}, which provides semantic or structural guidance (e.g., edited RGB frames, depth maps, or pose skeletons), and (2) a \textit{mask video}, which indicates per-frame regions to be preserved (black pixels) or regenerated (white pixels).

Our initial attempt adopts a straightforward multi-keyframe strategy: we construct the reference video by placing the user-edited keyframes at their corresponding timestamps, while filling all non-keyframe positions with a uniform gray frame (RGB value 127). Concurrently, the mask video is set to black at keyframes (instructing the model to faithfully reconstruct the edited content) and white elsewhere (prompting content generation based on the reference). However, this naive setup leads to severe temporal flickering and inconsistent motion in both the 1.3B and 14B parameter variants of VACE. We hypothesize that this instability arises because VACE was not trained on such sparse, discontinuous reference signals, resulting in hallucinated textures and incoherent motion interpolation between keyframes.

To mitigate this issue, we design a more stable reference configuration. Specifically, we retain the edited RGB image only at the first keyframe (typically the first frame of the video) in the reference video. For all subsequent keyframes, we replace the edited RGB content with the corresponding \textit{depth maps} derived from the original edited frames. Depth maps are chosen because they encode structural and motion cues with significantly less high-frequency appearance variation than RGB images, thereby reducing the risk of introducing inconsistent visual signals. Meanwhile, the mask video is simplified: only the first frame is set to black (to anchor the edited appearance), and all remaining frames (including other keyframes—are set to white), indicating that the model should synthesize these frames using the reference guidance.

This refined setup effectively decouples appearance and motion guidance: the model learns the target appearance from the first edited frame, while leveraging depth-derived structural cues from later keyframes to preserve the original video's motion dynamics. Under this configuration, VACE produces visibly more stable outputs with reduced flickering, enabling us to meaningfully analyze the limitations of sparse multi-keyframe editing

\section{Details of Training Pipelines}
\begin{algorithm}[h]
    \small 
    \caption{Source Fidelity Pipeline (Pseudo-Source Generation)}
    \label{alg:pseudo_source}
    \begin{algorithmic}[1]
        \Require Target video $\X = \{\boldsymbol{x}_t\}_{t=0}^T$, Filler video pool $\mathcal{P}$
        \Ensure Pseudo-source video $\Xt = \{\tilde{\boldsymbol{x}}_t\}_{t=0}^T$
        \State Sample filler video $\mathcal{Y} = \{\boldsymbol{y}_t\} \sim \mathcal{P}$
        \State Init mask shape $S \in \{\text{rect, ellipse, poly}\}$
        \State Init position $\boldsymbol{p}_0$, motion $\boldsymbol{v}$, rotation $\theta$
        \For{$t = 0$ to $T$}
        \State Update mask pose: $\boldsymbol{p}_t \leftarrow \boldsymbol{p}_{t-1} + \boldsymbol{v}, \quad \theta_t \leftarrow \theta_{t-1} + \Delta\theta$
        \State Handle boundary collisions and bounce $\boldsymbol{v}$ if needed
        \State Generate binary mask $\boldsymbol{m}_t$ from $S, \boldsymbol{p}_t, \theta_t$
        \State Extract patch from $\boldsymbol{y}_t$ (using ping-pong loop)
        \State $\tilde{\boldsymbol{x}}_t \leftarrow \boldsymbol{m}_t \odot \boldsymbol{y}_t + (\mathbf{1} - \boldsymbol{m}_t) \odot \boldsymbol{x}_t$
        \EndFor
        \State \Return $\Xt$
    \end{algorithmic}
\end{algorithm}

\begin{algorithm}[h]
    \small 
    \caption{Anchored Control Pipeline (Degraded Reference)}
    \label{alg:degraded_reference}
    \begin{algorithmic}[1]
        \Require Target video $\X$, Indices set $\{0, \dots, T\}$
        \Ensure Degraded reference video $\Xh$
        \State $\K \leftarrow \{0, T\}$ \Comment{Always include start/end}
        \State Sample $N$ random indices from $(0, T)$ into $\K$
        \State Sort $\K = \{k_0, k_1, \dots, k_{|\K|}\}$
        \For{each keyframe index $k \in \K$}
        \State $\hat{\boldsymbol{x}}_k \leftarrow \boldsymbol{x}_k$
        \If{random() $< 0.5$} \Comment{Geometric Degradation}
        \State Apply random Zoom-Stretch affine transform to $\hat{\boldsymbol{x}}_k$
        \EndIf
        \If{random() $< 0.5$} \Comment{Appearance Degradation}
        \State Define local region $\boldsymbol{b}_k$ (random blob)
        \State $\hat{\boldsymbol{x}}_k \leftarrow (\mathbf{1} - \boldsymbol{b}_k) \odot \hat{\boldsymbol{x}}_k + \boldsymbol{b}_k \odot \text{Blur}(\hat{\boldsymbol{x}}_k)$
        \EndIf
        \EndFor
        \For{$t = 0$ to $T$} \Comment{Temporal Interpolation}
        \State Find neighbors $k_i, k_{i+1} \in \K$ such that $k_i \le t \le k_{i+1}$
        \State $\alpha \leftarrow (t - k_i) / (k_{i+1} - k_i)$
        \State $\hat{\boldsymbol{x}}_t \leftarrow (1 - \alpha)\hat{\boldsymbol{x}}_{k_i} + \alpha \hat{\boldsymbol{x}}_{k_{i+1}}$
        \EndFor
        \State \Return $\Xh$
    \end{algorithmic}
\end{algorithm}

\begin{figure*}[p]
    \centering
    \includegraphics[width=0.9\linewidth]{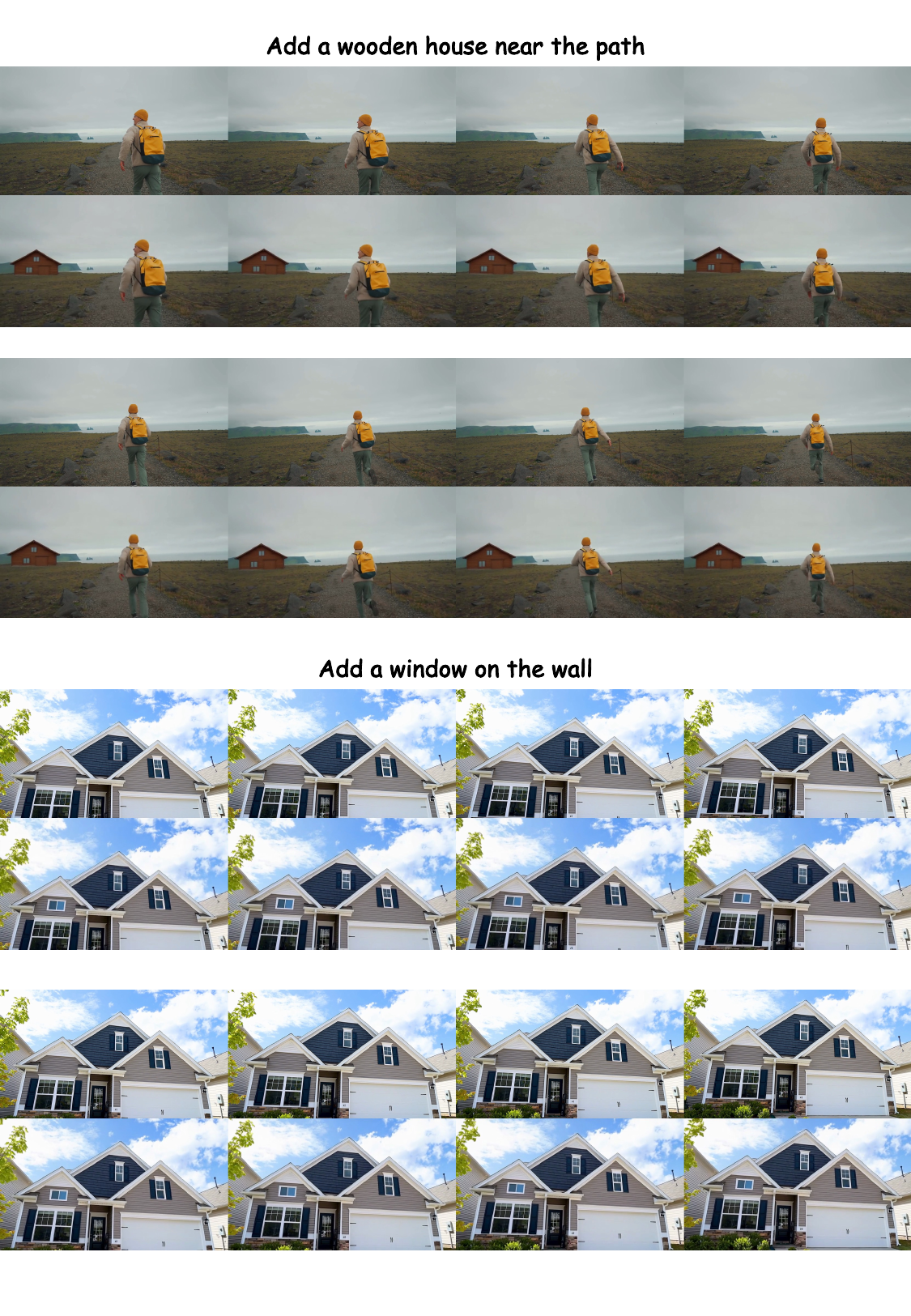}
    \caption{More visual results (1).}
    \label{fig:suppl_page1}
\end{figure*}

\begin{figure*}[p]
    \centering
    \includegraphics[width=0.9\linewidth]{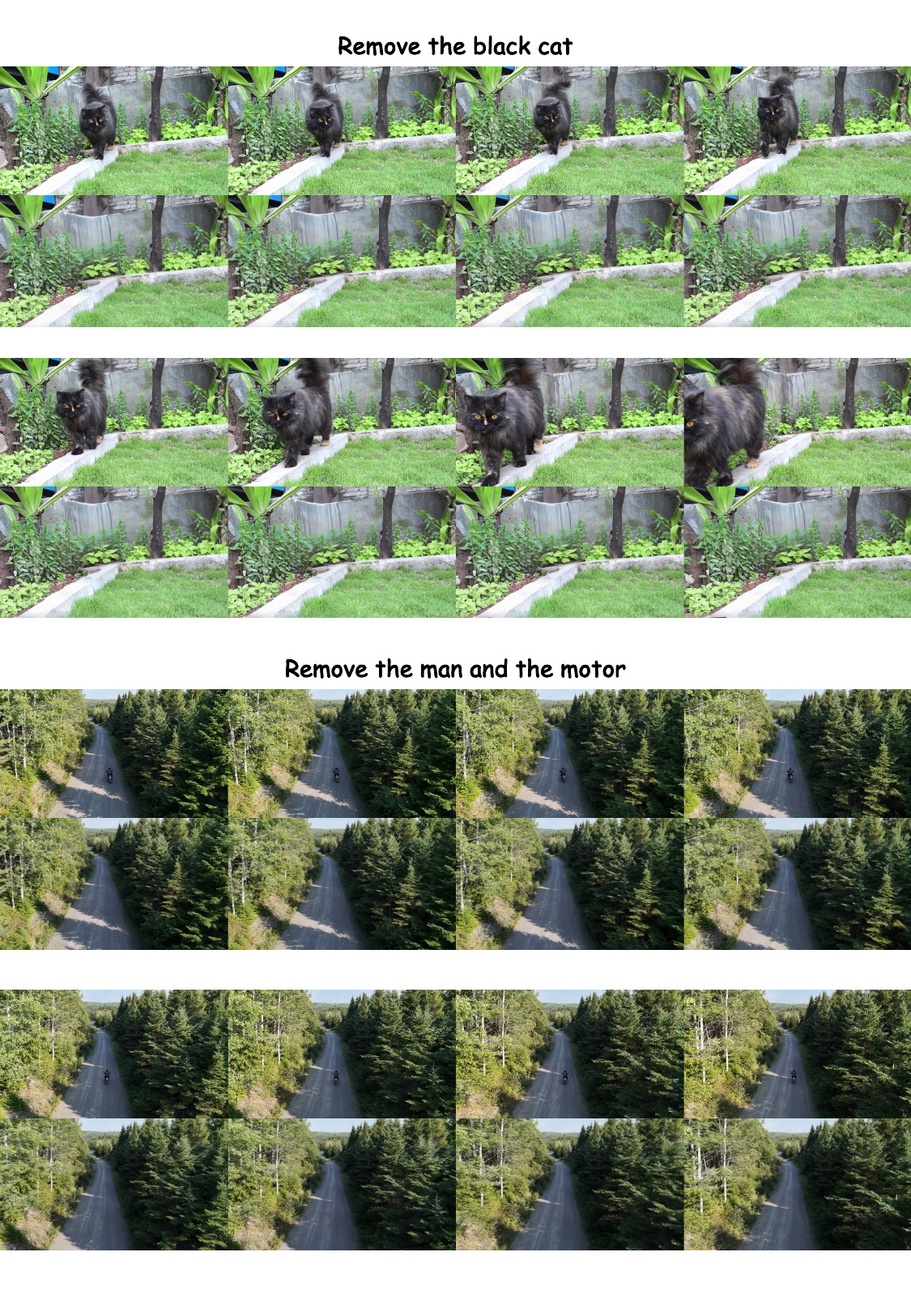}
    \caption{More visual results (2).}
    \label{fig:suppl_page2}
\end{figure*}

\begin{figure*}[p]
    \centering
    \includegraphics[width=0.9\linewidth]{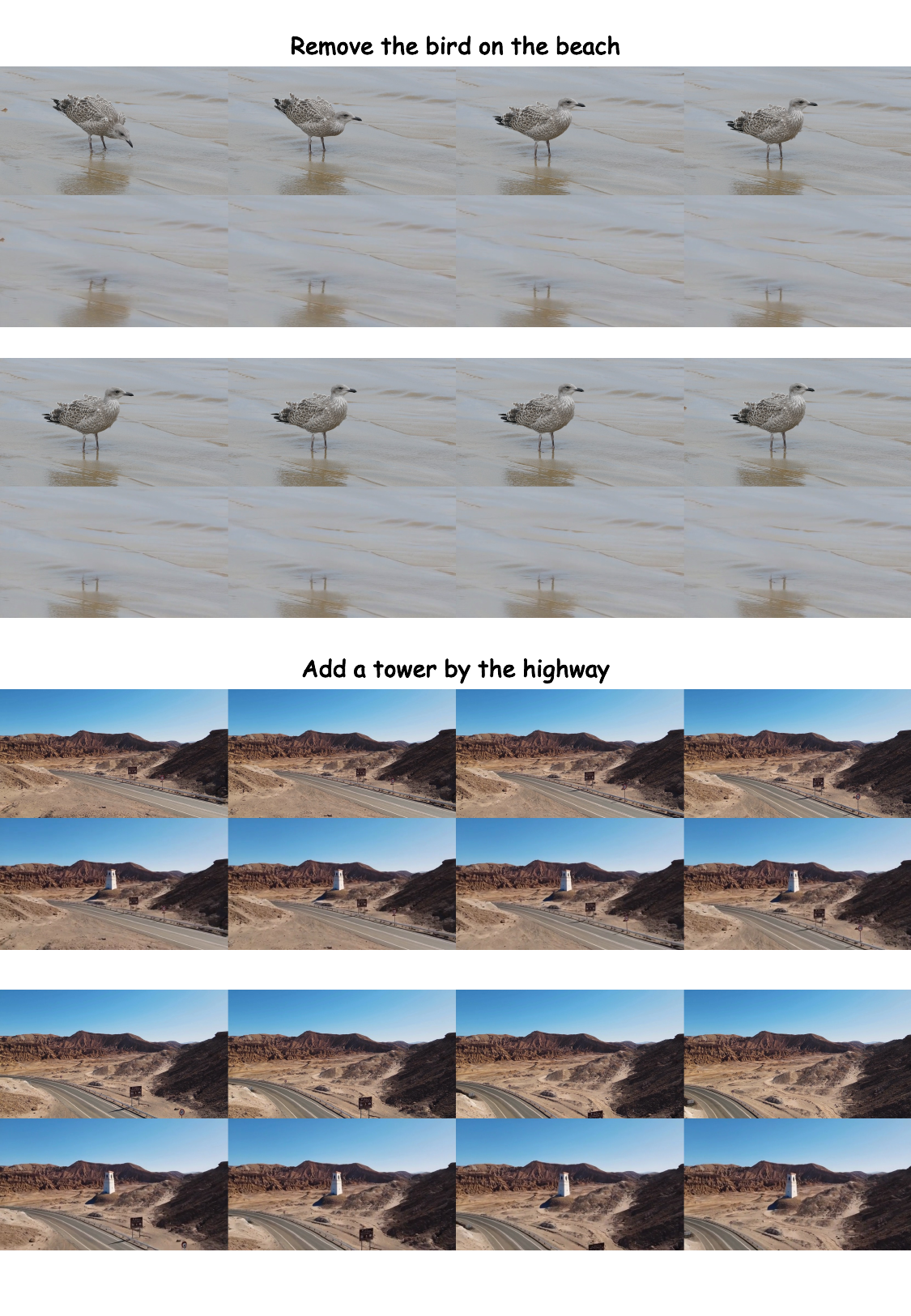}
    \caption{More visual results (3).}
    \label{fig:suppl_page3}
\end{figure*}

\begin{figure*}[p]
    \centering
    \includegraphics[width=0.9\linewidth]{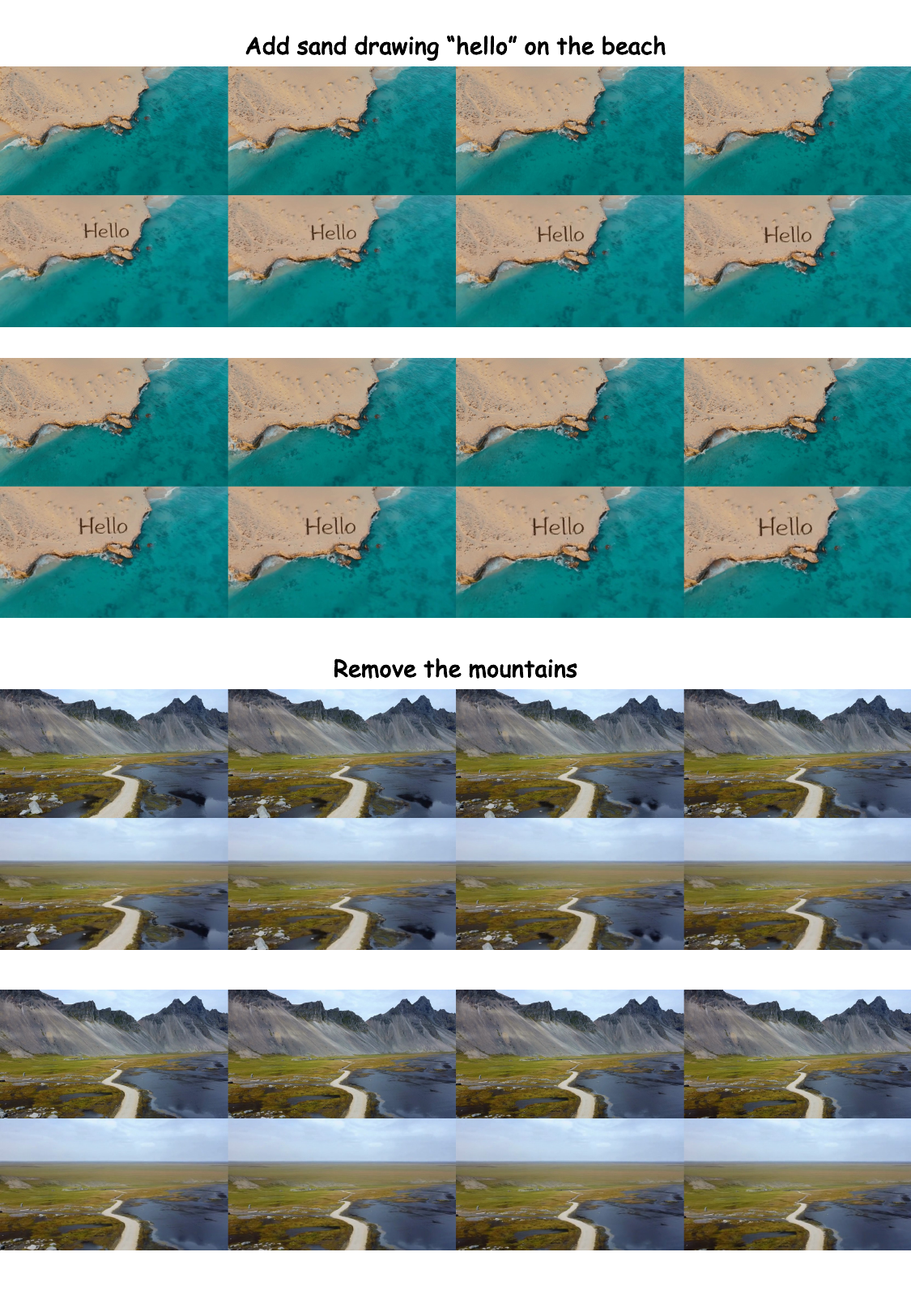}
    \caption{More visual results (4).}
    \label{fig:suppl_page4}
\end{figure*}

\end{document}